\documentclass[lettersize,journal]{IEEEtran}
\usepackage{amsmath,amssymb,amsfonts}
\usepackage{blindtext}
\usepackage[hidelinks]{hyperref}
\usepackage{graphicx}
\usepackage{textcomp}
\usepackage{xcolor}
\usepackage{graphics}
\usepackage{float}
\usepackage{orcidlink}
\usepackage{caption,subcaption}
\usepackage{placeins}
\usepackage{multirow}
\usepackage{tabularx,ragged2e}
\usepackage{booktabs}
\usepackage{multicol}
\usepackage{algorithm} 
\usepackage{algpseudocode}
\usepackage{xcolor}
\usepackage{listings} 
\usepackage{diagbox}
\usepackage{arydshln}
\usepackage{svg}
\usepackage[T1]{fontenc}
\lstdefinestyle{mystyle}{
    basicstyle=\ttfamily\footnotesize,
    breakatwhitespace=false,         
    breaklines=true,                 
    captionpos=b,                    
    keepspaces=false,                 
    numbers=left,                    
    numbersep=2pt,                  
    showspaces=false,                
    showstringspaces=false,
    showtabs=false,                  
    tabsize=2
}
\usepackage{arydshln}

\usepackage{tikz}
\usetikzlibrary{arrows.meta}
\usepackage{mathtools}
\newcommand\givenbase[1][]{\:#1\lvert\:}
\let\given\givenbase

\DeclarePairedDelimiterX\Basics[1](){\let\given\sgiven #1}
\newcommand\prob{P\Basics}
\newcommand\lgt{\text{logit}\Basics}
\newcommand\lgti{\text{logit}^{-1}\Basics}

\begin{document}

\title{Aligning Crowd-sourced Human Feedback for Reinforcement Learning on Code Generation by Large Language Models}

\author{Man Fai Wong, 
Chee Wei Tan 


\thanks{Man Fai Wong is with the Department of Computer Science, City University of Hong Kong, Hong Kong, China. (e-mail: mfwong29-c@my.cityu.edu.hk).}
\thanks{Chee Wei Tan is with the College of Computing and Data Science, Nanyang Technological University, Singapore. (e-mail: cheewei.tan@ntu.edu.sg).}
}



\maketitle

\begin{abstract}
This paper studies how AI-assisted programming and large language models (LLM) improve software developers' ability via AI tools (LLM agents) like Github Copilot and Amazon CodeWhisperer, while integrating human feedback to enhance reinforcement learning (RLHF) with crowd-sourced computation to enhance text-to-code generation. Additionally, we demonstrate that our Bayesian optimization framework supports AI alignment in code generation by distributing the feedback collection burden, highlighting the value of collecting human feedback of good quality. Our empirical evaluations demonstrate the efficacy of this approach, showcasing how LLM agents can be effectively trained for improved text-to-code generation. Our Bayesian optimization framework can be designed for general domain-specific languages, promoting the alignment of large language model capabilities with human feedback in AI-assisted programming for code generation.
\end{abstract}


\IEEEPARstart{D}ijkstra, in his seminal paper on computer-assisted programming \cite{dijkstra}, emphasized that while building large, complex programs is difficult, breaking down the process into smaller decisions makes it more manageable than relying solely on debugging. This approach not only prevents bugs but also aids in understanding the program by effectively reading its proof of correctness. Today, this concept of automated code generation is realized through Artificial Intelligence (AI) tools, which leverage Natural Language Processing (NLP) and Large Language Model (LLM) like OpenAI's GPT to comprehend, interpret, and engage with human language. This marks a significant evolution in AI-assisted programming \cite{dijkstra,MITcodegeneration,dijkstra1972humble}. By leveraging AI tools (LLM agents) like OpenAI ChatGPT~\cite{openai_2023}, Github Copilot~\cite{friedman2021introducing}, Copilot for Xcode \cite{tancopilot}, Google AlphaCode~\cite{li2022competition}, Amazon CodeWhisperer~\cite{amazon}, Codeium~\cite{codeium}, computer programmers worldwide now benefit from automated support to enhance efficiency and productivity in  software development \cite{wong2023natural}. 

Recent efforts in the software engineering community have embraced the notion that programming is an act of communication, known as the {\it software naturalness hypothesis} \cite{naturalness,naturalnessit,naturalnessbuggy,bigcodesurvey}. This approach involves analyzing programming and natural languages by developing probabilistic models of source code and leveraging the patterns in coding, particularly in the context of large codebases. The logical patterns present in numerous code snippets can be leveraged to understand how programmers read and internalize the logic flow of computer source code. For instance, how do novice programmers interpret a simple {\tt hello world} program compared to how experienced programmers analyze a code snippet on algorithms? With the growing availability of extensive data, a key question arises: How can we effectively leverage unsupervised machine learning in combination with the naturalness hypothesis, while integrating human feedback through human-in-the-loop computation, to enhance and evaluate LLMs and LLM agents?

Human-in-the-loop computation has the potential to greatly improve the readability, reliability, and testability of generated code, while effectively addressing AI alignment issues in code generation \cite{von2004telling, vonahncaptcha, von2007improving, von2004labeling}. The AI alignment challenge in LLM-based code generation involves ensuring that the generated code meets requirements and is executable. LLMs may prioritize fluency over accuracy, producing code that compiles but does not function as intended. Also, understanding and adapting to developer preferences remains a challenge. Efforts to enhance code alignment in LLMs emphasize combining reinforcement learning, human-assisted computation, and fine-tuning techniques. A generalized training approach may utilize prompted language models and curated data, balancing speed and accuracy to improve code generation precision and reliability.

\begin{figure*}[t!]
    \centering
    \includegraphics[width=\linewidth]{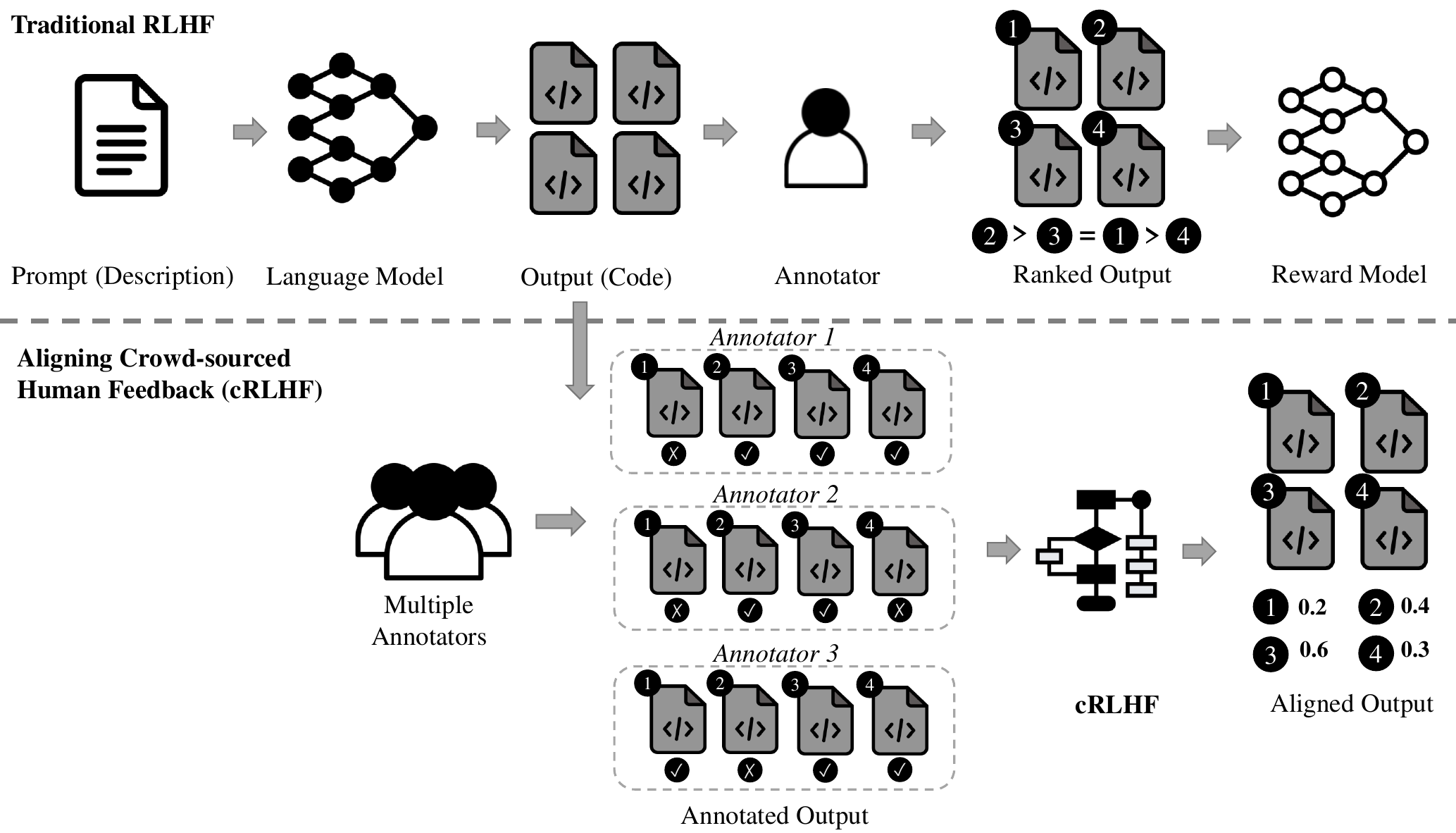}
    \caption{The schematic diagram of traditional RLHF and aligning crowd-sourced human feedback strategy on code generation. The upper part of the Figure shows the traditional RLHF method, which employs an annotator to rank the generated outputs and forward them to the reward model. The lower part presents how crowd-sourced RLHF strategy can be done with multiple annotators, which automatically compute the consensual ranked output in terms of ranking or reward scores.}
    \label{fig:overview}
\end{figure*}

LLMs are enabled by Reinforcement Learning with Human Feedback (RLHF) \cite{christiano2017deep, stiennon2020learning, ouyang2022training}, which has recently advanced unsupervised learning for complex tasks. RLHF utilizes human evaluations to direct the training of RL models, enhancing language models to produce outputs that align with human expectations. By converting human assessments into goal-oriented feedback, RLHF has facilitated the development of large language models like OpenAI Codex and GPT, allowing them to tackle a broader range of tasks. However, the effectiveness of raw human feedback as reward signals diminishes in uncertain environments, leading to the exploration of new strategies to minimize feedback requirements.

In this paper, we present a framework, cRLHF, for aligning crowd-sourced human feedback from multiple sources. This approach seeks to improve and simplify the conventional human ranking process in RLHF systems by computing reward scores without requiring the training of an additional reward model. As shown in Figure \ref{fig:overview}, the framework is specifically applied to language-to-code generation tasks. In our paper, we build upon our prior work in \cite{ling2018human,rlhfbayesian0} using Bayesian analysis as a novel approach to policy optimization, offering a probabilistic framework that enhances decision-making by incorporating uncertainty and prior knowledge. Our proposed framework has been designed to cater to the self-assessment needs of participants who submit their annotations on LLM-generated outputs. This is particularly relevant when limited prior knowledge complicates locating error-prone code lines. Our framework involves designing a statistical learning and validation process where all human annotators receive the same input (e.g., code snippet), evaluate it independently, and contribute to an aggregated output (e.g., an intrinsic attribute of the code). We conduct experiments on code generation by further fine-tuning baselines and assessing benchmark performance.

In summary, the contributions of our work are as follows:
\begin{itemize}
    \item We propose a novel framework for aligning crowd-sourced human feedback, named cRLHF, which leverages RLHF mechanism from multiple annotators based on Bayesian inference, and aligns the ranking from different sources and computes the reward score without additional reward modeling.
    
    \item We develop a Bayesian learning algorithm with an optimization perspective based on regularized logistic regression, aimed at improving accuracy and reducing uncertainty in crowd-sourced code annotation.

    \item We evaluate our proposed framework on an established benchmark with a comprehensive set of experiments, involving annotators to assess the generated outputs. In this work, we include a range of more recent and larger baseline models that are specifically pre-trained on code. Our results show that our framework improves the performance of these baseline models, generating higher-quality code and better results across the benchmarks.
\end{itemize} 

The rest of this the paper is organized as follows: Section \ref{sec:works} provides an introduction to related works within the field. Section \ref{sec:crlhf} describes our proposed framework for aligning crowd-sourced human feedback on the RL system. Section \ref{sec:experiment}  details the experimental configurations and baselines employed in this study. Section \ref{sec:results} presents the results and a comprehensive discussion. Finally, Section \ref{sec:conclusion} serves as the conclusion, encapsulating the key insights and contributions of this paper.

\section{Related Works}\label{sec:works}
\subsection{Large Language Models for Code Generation}
Text-to-code generation, also known as program synthesis and code generation, is the process of converting natural language text, typically in the form of human-readable descriptions or instructions, into executable code in a programming language \cite{waldinger1969prow}. Conversely, code-to-code generation encompasses methods like code translation and code completion \cite{wong2023natural}. Prior approaches for generating source code from natural language viewed the task as an encoder-decoder learning endeavor, as demonstrated by models like code2seq \cite{alon2018code2seq} and code2vec \cite{alon2019code2vec}. The Seq2AST model \cite{rabinovich2017abstract} introduced an Abstract Syntax Tree (AST)-based decoder for code generation, while a syntactic neural model \cite{yin2017syntactic} incorporated an AST-based decoder and a grammar model to account for the underlying syntax of the target programming language.

More recently, transformer-based LLMs through attention mechanism \cite{vaswani2017attention} have showcased remarkable prowess in code generation tasks \cite{raffel2020exploring, brown2020language, lewis2020bart}. These models utilize contextual representations extracted from large amounts of code known as Big Code \cite{vechev2016programming}, in addition to publicly available code sources and natural language data, to amplify program synthesis. Also, pre-trained LLMs with extensive parameters dedicated to code generation \cite{feng2020codebert, fried2022incoder, le2022coderl, zheng2023codegeex}, have been trained through expansive datasets. Also, these models have been fine-tuned for specific downstream tasks, encompassing code-to-code generation by the source code with corresponding problem descriptions. These models can generate new code snippets or entire programs in a designated programming language. 

However, it is imperative to acknowledge that while language models trained on code can produce unreliable and legible code, there exists no assurance that the generated code will be accurate or optimal \cite{fan2022improving}. Hence, circumspection must be exercised when incorporating generated code, and exhaustive testing should be undertaken prior to deploying it within a production environment. 

\subsection{Reinforcement Learning with Human Feedback} \label{sec:hrl}
Code generation can be framed as a sequence generation task, which is well-suited for optimization through reinforcement learning (RL) methods \cite{ranzato2015sequence}. RL, an effective unsupervised learning technique, improves policies by leveraging a feedback loop driven by rewards, thus enhancing decision-making in various states. This approach is particularly effective for fine-tuning large language models (LLMs) in conditional text generation tasks \cite{le2022coderl, chen2018executionguided, shojaee2023executionbased}. By applying RL, we can significantly enhance code generation across different sizes of pre-trained models. This strategy involves training an RL agent to generate code, refining its outputs by maximizing a reward function based on code quality and correctness.

Recent advancements have focused on incorporating human feedback into the learning process \cite{christiano2017deep, stiennon2020learning, ouyang2022training, bai2022training, rafailov2024direct, hou2024chatglm}, allowing models to be guided by human preferences rather than relying solely on environment-based rewards. Human feedback, often collected through interactions with dialogue agents \cite{glaese2022improving, openai_2023}, plays a crucial role in evaluating generated outputs. This feedback enhances the model's ability to align its decisions more closely with human expectations, improving the overall performance in RL-based systems.

Despite the progress in RLHF-based approaches, several limitations persist. These methods typically require a significant amount of human feedback for model fine-tuning, which can be both time-consuming and costly. Moreover, the quality and consistency of human feedback can vary, posing challenges for effective model training. Recent studies have aimed to address these issues. For instance, \cite{dong2023raft} introduced a new alignment algorithm within the alignment pipeline for generative models, extending beyond LLMs. Similarly, \cite{chen2023improving} proposed incorporating natural language feedback into the RL process to enhance code generation. Additionally, \cite{ramamurthy2023reinforcement} presented a novel policy optimization algorithm as an alternative to policy gradient methods, designed to better align LLMs with human preferences. The work in \cite{rlhfbayesian1} explored the interpretation of RL with Kullback-Leibler penalties as a form of Bayesian inference, while \cite{rlhfbayesian2} proposed a Bayesian framework to model the distribution of disagreements in human preferences during the training of a preference model.

\begin{figure}[t!]
    \centering
    \includegraphics[width=\linewidth]{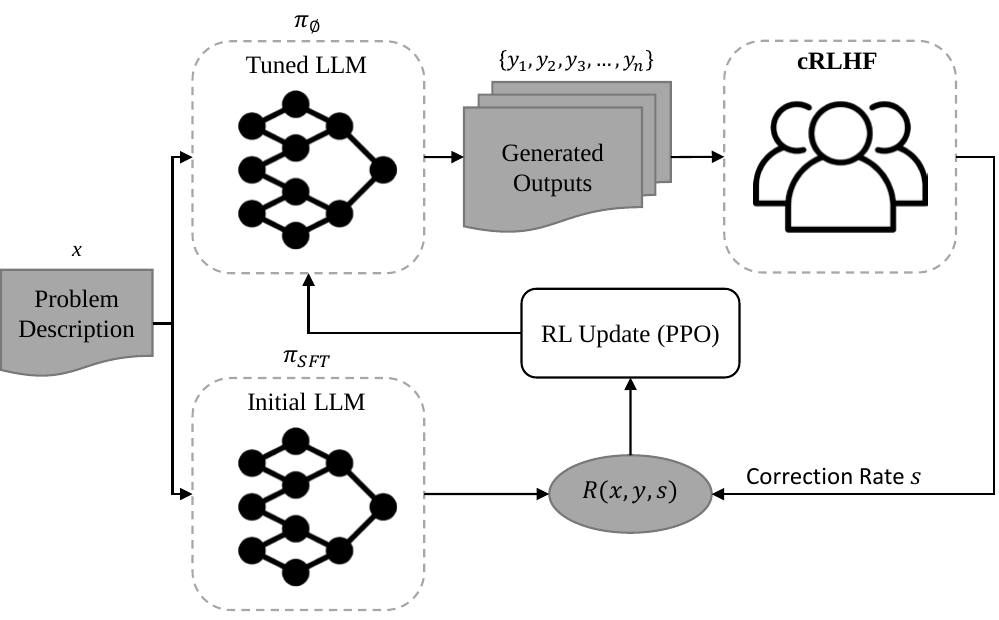}
    \caption{An overview of the cRLHF framework with the reinforcement learning algorithm utilizing proximal policy optimization (PPO) for code generation.}
    \label{fig:workflow}
\end{figure}

\begin{figure*}[h!]
    \centering
    \includegraphics[width=\linewidth]{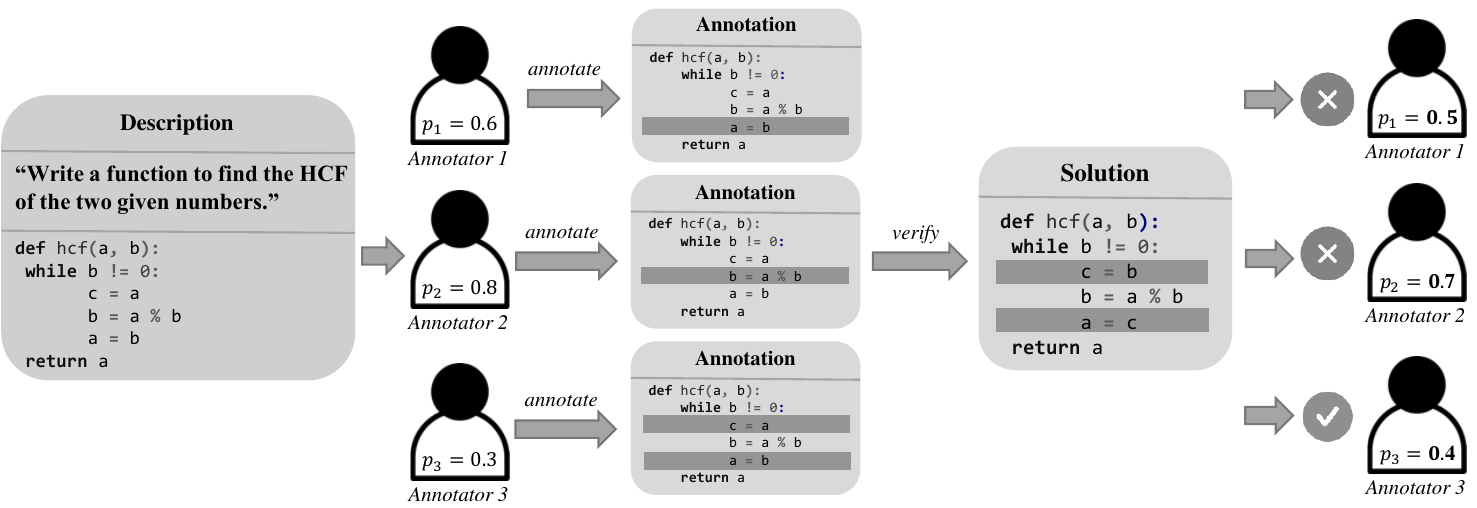}
    \caption{The overview of the process for evaluating $p_i$ across all annotators. Presented with descriptions and code examples, as depicted on the left-hand side of the figure, each annotator, with their respective $p_i$ values, is prompted to identify errors within the code. The system then assesses the accuracy of the annotations and proceeds to update the $p_i$ values accordingly.}
    \label{fig:pi}
\end{figure*}

\section{Methodology} \label{sec:crlhf}
\subsection{Problem Description}\label{sec:problem}
To start, we present a formal description of the code generation problem using LLMs. Our approach primarily follows the framework established in \cite{ziegler2019finetuning} for integrating human feedback into LLM training. The workflow of our conditional reinforcement learning from human feedback (cRLHF) framework is illustrated in Figure \ref{fig:workflow}. In our proposed framework, a programming problem description $x \in X $ serves as the input prompt. The supervised fine-tuned (SFT) model $\pi_{\text{SFT}}$ is designed to translate this description from natural language into a corresponding code snippet or program $y$. A reward function $R(x, y, s)$ evaluates the generated output, with the reward score $s$ calculated through our crowd-sourcing algorithm based on human feedback. Our goal is to fine-tune a model $\pi_\theta$, which initially corresponds to $\pi_{\text{SFT}}$, to generate responses that maximize these rewards.

\subsection{Aligning Crowd-sourced Human Feedback for Code Generation}
\label{sec:hacm}
Let us describe the cRLHF framework, which aligns crowd-sourced human feedback by computing the reward function $s$ using input from multiple annotators. For input $\{x_1,x_2,\dots,x_q,\dots,x_t\}\in X$, the collection of all generated code snippets from $\pi_\theta$ can be represented as $(y_1,y_2,y_3,\dots,y_n)$ of $y \in Y$, where $n$ signifies the count of generated outputs. For the input $\{x_1,x_2,\dots,x_q\}$, we know the exact correctness for each generated output. Our human feedback is procured from a group of $m$ annotators, denoted as $(a_1,a_2,a_3,\dots,a_m)$ of $a \in A$. These annotators then are tasked with providing annotations for each $y$. In this work, our annotation methodology revolves around evaluating each line $l$ of the program, wherein $y = \{ l_1, l_2, l_3, \dots, l_k\}$ with $k$ denoting the number of lines. Workers annotate each line of code using labels such as ``correct'' for error-free code or ``wrong'' for buggy code, which are represented by values $\{1,-1\}$. For each line of code, it has a prior probability of being correct and the prior probability is $\frac{1}{2}$. For $l_i$, the correctness is represented by $L_0$, where
\begin{align}
    L_0 = 
    \begin{cases}
    1 & \text{if the $l_i$ is true}\\
    -1 & \text{otherwise.}
    \end{cases}
\end{align}
We begin with the assumption that the system has no prior knowledge of the value of $L_0$, and its task is to compute the likelihood of $l_i$ being true ($\prob{L_0 = 1}$) based on human annotator input. Let the prior probability of $l_i$ being true be $p_0$, e.g., $p_0 = \prob{L_0 = 1} = \frac{1}{2}$.

Consider the response from an annotator $a_1$ as $L_1$, with a corresponding probability of correctness denoted as $p_1$, namely:

$$
p_1 = \prob{L_1 = 1 \given L_0 = 1} = \prob{L_1 = -1 \given L_0 = -1}.
$$

From Bayes' Theorem, we have:

$$
\prob{L_0 = 1 \given L_1 = 1} = \frac{\prob{L_1 = 1 \given L_0 = 1}\prob{L_0 = 1}}{\prob{L_1 = 1}}.
$$

We can further rewrite it using $p_0$ and $p_1$:

$$
\prob{L_0 = 1 \given L_1 = 1} = \frac{p_1 p_0}{p_1 p_0 + (1 - p_1)(1 - p_0)}.
$$

By introducing the logit function \cite{jurafsky2000speech}, we can further simplify the probability into:
\begin{equation}
    \prob{L_0 = 1 \given L_1 = 1} = \lgti{\lgt{p_0} + \lgt{p_1}},
    \label{eq:1}
\end{equation}

where the logit function is defined as $\lgt{\cdot} = \log(\frac{\cdot}{1 - \cdot})$, and $\lgti{\cdot} = \frac{\exp{\cdot}}{1 + \exp{\cdot}}$. Similarly we have:
\begin{equation}
    \prob{L_0 = 1 \given L_1 = -1} = \lgti{\lgt{p_0} - \lgt{p_1}}.
    \label{eq:-1}
\end{equation}

Combining \eqref{eq:1} and \eqref{eq:-1}, we get: 
\begin{align}
    \prob{L_0 = 1 \given L_1 = \epsilon_1} = \lgti{\lgt{p_0} + \epsilon_1 \lgt{p_1}}.
    \label{eq:a1}
\end{align}

Given that the available options for annotators are restricted to $\{1,-1\}$ within this system, it follows that $\lgt{p_0} = 0$. With this insight, we can simplify \eqref{eq:a1} to:
\begin{align}
    \prob{L_0 = 1 \given L_1 = \epsilon_1} = \lgti{\epsilon_1 \lgt{p_1}}.
    \label{eq:a2}
\end{align}

We then consider another annotator $a_2$ contributes their response as $L_2$ with an independent probability $p_2$ of being correct, we can regard the posterior probability $P(L_0=1|L_1=\epsilon_1)$ as our revised prior. Then we obtain:
\begin{align}
    &\prob{L_0 = 1 \given L_1 = \epsilon_1, L_2 = \epsilon_2} \nonumber \\
    =\ &\lgti{\lgt{P(L_0=1|L_1=\epsilon_1} + \epsilon_2\lgt{p_2} } \nonumber \\
    =\ &\lgti{\epsilon_1\lgt{p_1} + \epsilon_2\lgt{p_2}} \label{eq:an}.
\end{align}

Building upon the analysis involving two annotators as discussed earlier, we now extend our consideration to a broader scenario where the system has acquired feedback from $n$ annotators regarding $l_1$:
\begin{align}
    &\prob{L_0 = 1 \given L_1 = \epsilon_1, \cdots, L_n = \epsilon_n} \nonumber \\
    =\ & \lgti{\sum_{i = 1}^n \epsilon_i \lgt{p_i}} \label{eq:an2}.
\end{align}

We extend our previous work in \cite{ling2018human,rlhfbayesian0} to establish a systematic procedure for evaluating the $p_i$ values as illustrated in \ref{fig:pi}. To determine the $p_i$ value associated with each annotator, we initiate the $p_i$ value with a parameter denoted as $\nu $. For the initial set of $q$ annotation tasks the system is furnished with correct answers from the original dataset $X$ which can be consider as the Honeypot questions, enabling it to utilize these problem sets and annotator responses to iteratively refine each annotator's $p_i$ values. This fine-tuning process entails adjusting the annotators' $p_i$ values, either upwards or downwards, with a constant step size, depending the accuracy of their annotations for $q$ tasks.  The $\bar p$ value modulates the adjustment, ensuring it aligns with the system's confidence in its response. After annotating each program in the $p$ tasks, the system updates each annotator's $p_i$ accordingly:
\begin{equation}
    p^*_{i} = \lgti{\lgt{p_i} + \lambda \mu \ \lgt{\bar p}},
    \label{eq:p-update}
\end{equation}
where $p^*$ is the updated prior probability value, $\mu$ is the correctness of the annotator's response to the particular $l_i$ ($\mu = 1$ for correct annotation, $\mu = -1$ for incorrect response), $\lambda$ is a hyper-parameter, and $\bar p$ is the certainty that the response used by the system for annotator is correct. For the initial questions and the evaluation questions, $\bar p$ would be $1$, meaning that the system is absolutely sure about the answer. For other questions, $\bar p$ would be the value obtained from \eqref{eq:an}.

Referring back to our generated outputs, denoted as $y$ and comprising a set of possible lines of code $\{l_1, l_2, l_3, \dots , l_k\}$, where $k$ represents the total number of lines, we can assess the accuracy of each line using the equation \eqref{eq:an2}. The resulting count, denoted as $c$, represents the number of correct lines across all the lines evaluated. This count then serves as the basis for calculating the \textit{correction rate} on a line-to-line basis. This correction rate, termed the aligned score and represented as $s=c/k$, which $s \in [0,1]$ for each $y$ within the system. We then can integrate this score into the system without relying on reward modeling, similar to existing preference-based {RL} approaches \cite{hejna2024inverse,an2024direct}.Subsequently, we construct a dataset $S$, consisting of triplets $(x, y, s)$ that capture annotators' feedback for all generated output of $y$ to ranking score directly which is to map a given description and its corresponding code to a reward value $s$. Once the system computes the $s$ scores for each generated output based on the annotators' feedback, it employs the Proximal Policy Optimization (PPO) algorithm, as outlined by \cite{ziegler2019finetuning} and \cite{schulman2017proximal}, to optimize the learning process. To demonstrate our cRLHF framework (see Figure \ref{fig:workflow}), we outline the crowdsourcing computation and learning process in Algorithm \ref{alg:cRLHF}.

\begin{algorithm}
\caption{cRLHF with LLM for Code Generation} \label{alg:cRLHF}
\hspace*{\algorithmicindent} 

\textbf{Input:} Dataset $X$, SFT Model $\pi_{SFT}$

\textbf{Output:} Tuned Model $\pi_{\theta}$

\textbf{Parameters:} Correctness Threshold $\tau$
\begin{algorithmic}[1]
\State $\pi_{\theta} \leftarrow \pi_{SFT}$
\For{\text{Problem Input $\{x_1,x_2,\dots,x_q,\dots,x_k\}\in X$}}
        \If{Honeypot Questions $\{x_1,x_2,\dots,x_q\}$}
            \For{\text{Annotators $a\in A$}}
                \State $\{p\} \leftarrow$ Update by eq. (\ref{eq:p-update})
            \EndFor
        \Else
            \State $\{y_1,y_2,y_3,\dots,y_n \in Y\} \leftarrow
            \pi_{\theta}(x)$
            \For{\text{Generated Output $y\in Y$}}
            \State $c,k \leftarrow 0$
                \For{\text{Annotators $a\in A$}}
                    \State $\{\epsilon\} \leftarrow \text{Annotate}(y)$
                \EndFor
                \For{\text{Each Line $l \in y$}}
                    \State $k \leftarrow k+1$
                    \If{ eq. (\ref{eq:an2}) $> \tau$}
                        \State $c \leftarrow c+1$
                    \EndIf
                \EndFor
                \State $s \leftarrow c/k$
                 \For{\text{Annotators $a\in A$}}
                    \State $\{p\} \leftarrow$ Update by eq. (\ref{eq:p-update})
                \EndFor
            \EndFor
            \State $\pi_\theta \leftarrow \text{PPO}(x, y, s)$
        \EndIf
\EndFor
\end{algorithmic}
\end{algorithm}

\subsection{An Optimization Perspective}
This section introduces an optimization-based approach to assess annotators' confidence levels, contrasting with the iterative updates above. Rather than incremental adjustments, this allows a {\it one-shot update} that leverages all data at once to achieve a globally optimal confidence estimate. Here, language models are framed as log-linear models where features capture historical context up to a specified length, with parameters optimized through maximum likelihood. Though only a perspective, this optimization view scales linearly with corpus length and feature count, offering an efficient, stable alternative to sequential updates that can sometimes lead to drift and suboptimal confidence estimates \cite{chen2000survey,goodman2004exponential,NoahSmith2011,nelakanti2013structured}. This optimization perspective also offers a probabilistic interpretation akin to regularized logistic regression that provides fundamental performance bounds to the learning process of our cRLHF framework.

To enhance robustness and reduce overfitting in noisy annotations, we use sparsity regularization to align model predictions with a key subset of annotators’ responses, minimizing small weights to improve generalization by introducing a regularization term, $R(\cdot) = \Vert \cdot \Vert_1$ to the optimization problem formulation \cite{goodman2004exponential,NoahSmith2011,nelakanti2013structured}. Sparse regularization captures the idea that the utility of features in a predictive model is best understood collectively. It aims to identify compact and reliable groups of users in the crowd in aligning with the software naturalness hypothesis.

First, we perform a variable transformation, setting $\tilde{p}_i = \text{logit}(p_i)$ for each $i$ and define the vector $\mathbf{\tilde{p}} = (\tilde{p}_1,\dots,\tilde{p}_n)^T$ and $\boldsymbol{\epsilon} = (\epsilon_1,\dots,\epsilon_n)^T$. Next, substitute $\mathbf{\tilde{p}}$ into the likelihood function in (\ref{eq:an2}) and take its logarithm, yielding $\log \left( 1 + \exp \left( -\sum_{i = 1}^n \epsilon_i \tilde{p}_i \right) \right)$, which we denote by $\mathcal{L}(\mathbf{\tilde{p}})$. Consider the following convex objective function to minimize:
$$
\mathcal{L}(\mathbf{\tilde{p}}) + \gamma \sum_{i=1}^{n} | \tilde{p}_i |
= \log (1+ \exp( -\boldsymbol{\epsilon}^T \mathbf{\tilde{p}})) + \gamma \Vert \mathbf{\tilde{p}} \Vert_1,
$$
where $\gamma > 0$ is a regularization parameter that controls the strength of the regularization. In particular, $\nabla \mathcal{L}(\mathbf{\tilde{p}})$ is the gradient of the loss function $\mathcal{L}(\mathbf{\tilde{p}})$ with respect to $\mathbf{\tilde{p}}$, given by 
$\mathcal{L}(\mathbf{\tilde{p}}) = -\exp(-\boldsymbol{\epsilon}^T \mathbf{\tilde{p}}) \cdot \boldsymbol{\epsilon} / (1 + \exp(-\boldsymbol{\epsilon}^T \mathbf{\tilde{p}}))$, or in a concise form as 
$$
\nabla \mathcal{L}(\mathbf{\tilde{p}}) = -\lgti{ -\boldsymbol{\epsilon}^T \mathbf{\tilde{p}} } \cdot \boldsymbol{\epsilon}.
$$

To solve the above optimization problem, we use proximal gradient descent method, which handles the composite objective function by separating the smooth loss function from the regularization term \cite{parikh2014proximal}. The proximal gradient update rule is given by:
$$
\mathbf{\tilde{p}}^{(k+1)} = \text{prox}_{\eta \gamma R} \left( \mathbf{\tilde{p}}^{(k)} - \eta \nabla \mathcal{L}( \mathbf{\tilde{p}}^{(k)}) \right),
$$
or equivalently as the following proximal algorithm:
$$
\mathbf{\tilde{p}}^{k+1} = \arg \min_v \left\{ \frac{1}{2 \mu} \Vert  \mathbf{\tilde{p}}^{k} - \mu \nabla \mathcal{L}(\mathbf{\tilde{p}}^{(k)})  - v \Vert_2^2 + \eta_k \Vert v \Vert_1 \right\}
$$
where $\eta_k$ is a constant positive step size. Each iteration involves two proximal operators, similar to alternating projections: a gradient computation and a shrinkage operator, achieving an \(O(1/\kappa^2)\) iteration bound, where \(\kappa\) is a specified small error threshold. It requires a step size $\eta_k$ within $(0, 2/L)$, where $L$ is the Lipschitz constant of $\nabla \mathcal{L}(\mathbf{\tilde{p}}^{k})$.

The Lagrange dual problem for regularized logistic regression can be expressed as a single-variable convex optimization problem:
$$
\max_{0 < \nu < 1} - \nu \log  \nu  - (1-\nu) \log \left( 1 - \nu \right) \; \text{subject to} \; \|\epsilon\|_{\infty} \nu \leq \gamma,
$$
which is an infinity-norm constrained binary entropy maximization problem. To derive this, we introduce an auxiliary variable $u = \epsilon^T \tilde{p}$ and form the Lagrangian: $
\mathcal{L}(\tilde{p}, u, \nu) = \log(1 + \exp(-u)) + \gamma \|\tilde{p}\|_1 + \nu (u - \epsilon^T \tilde{p})$. Minimizing with respect to $u$ and $\mathbf{\tilde{p}}$ yields the dual problem with the constraint $\|\epsilon\|_{\infty} \nu \leq \gamma$, stemming from the dual norm of the $\ell_1$-norm regularization term, ensuring that the dual problem is bounded. This completes the derivation. 

Solving this dual problem provides a necessary lower bound on the costs associated with learning in a crowd-based setting. Specifically, since
$\tilde{\mathbf{p}}^{\star} = \text{logit}(\mathbf{p}^{\star})$, the optimal margin satisfies
$$
-\boldsymbol{\epsilon}^T \text{logit}(\mathbf{p}^{\star}) = \text{logit}(\nu^{\star}) \le \gamma/\|\epsilon\|_{\infty},
$$
where $\nu^{\star}$ is the optimal dual solution, which can be interpreted as a form of weighted averaging the logits of the data. In addition, this Lagrange dual approach, along with an iterative optimization algorithm, can guide initializations, such as setting $\bar{p}$ empirically in Bayesian learning. This initialization can serve as a fine-tuning layer in a neural network-based language model, improving prediction accuracy while ensuring well-regularized estimates of annotator reliability, an area for further exploration.

\section{Experimental Evaluations}\label{sec:experiment}
In this section, we present a comprehensive evaluation of the cRLHF framework by detailing the experimental approach, feedback collection process, evaluation benchmarks, and baseline models.
\begin{figure}[!t]
    \centering
    \includegraphics[width=\linewidth]{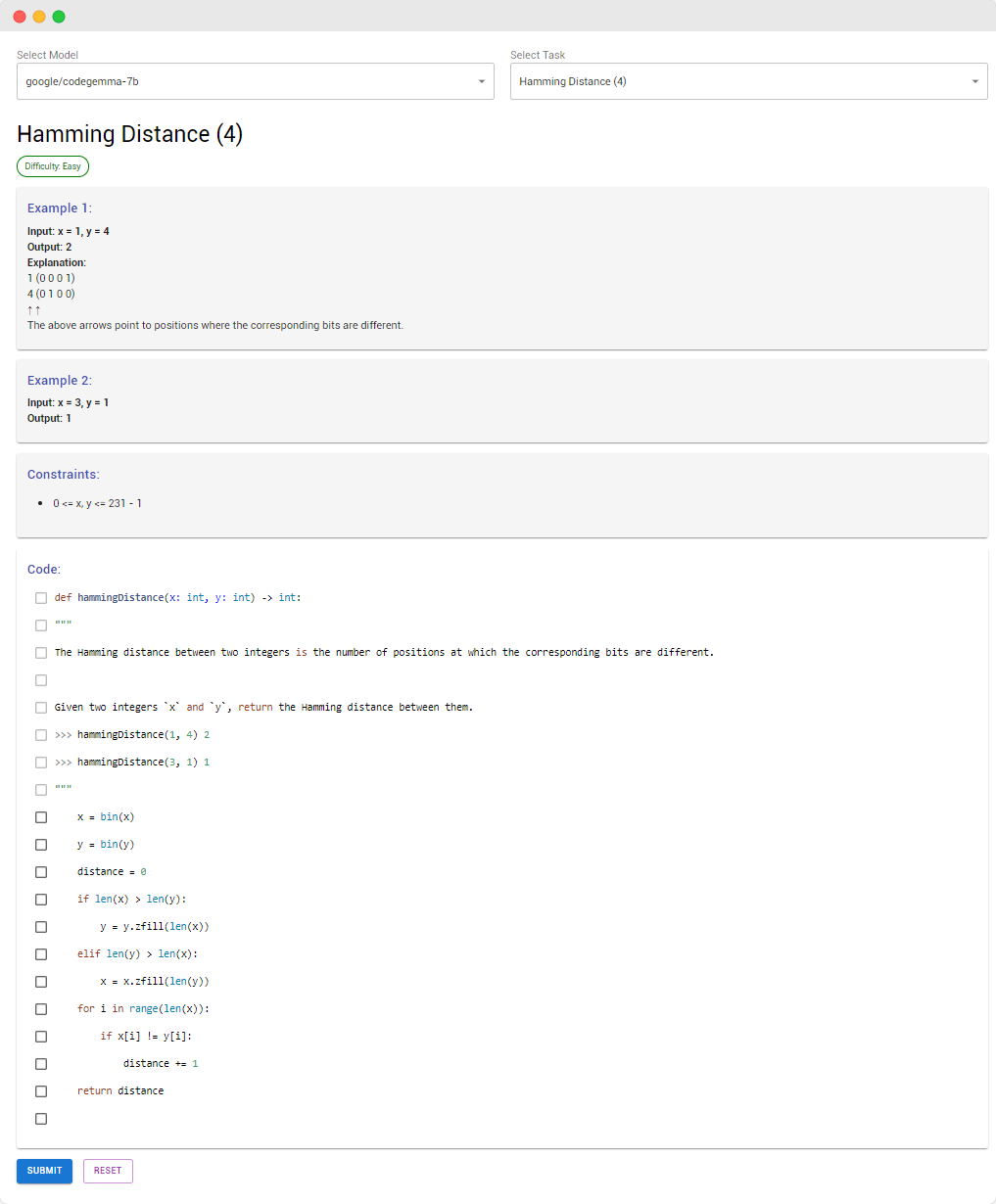}
    \caption{The user interface on online crowdsourcing platform for code annotation tasks. Annotators will receive various code snippets generated by LLMs alongside their corresponding descriptions. Their task involves annotating lines of the program that contain errors. A detailed guide on how to use this code annotation tool is also provided for annotators.}
    \label{fig:ui}
\end{figure}

\subsection{Evaluation Setup}\label{sec:setup}
Our experimental evaluation is designed to demonstrate the effectiveness of the cRLHF framework in improving code generation performance. All experiments are conducted using two NVIDIA A100-SXM-80GB GPUs and an AMD EPYC 7J13 64-Core Processor. The process begins by evaluating baseline models on code generation benchmarks, which serve as a reference for assessing their initial performance. Our evaluation follows the setting of the Bigcode evaluation framework \cite{bigcode-evaluation-harness} for code generation.

We then select a diverse set of code generation problems from external datasets that are not included in the original benchmarks. Code samples are generated for these problems using the baseline models. Human annotators evaluate the quality of the generated code and provide feedback. This feedback is processed by our cRLHF framework, which aligns and aggregates it into coherent scores for each output. The code generated by each model is produced independently, without any information being shared or mixed with other models during the generation process. For the parameters in the cRLHF process, we set $\lambda = 1$ and $\tau = 0.5$ in our experiments. Each model, using a temperature of $t = 0.8$ and a maximum output length of 128 tokens, generates its own code samples independently. The evaluation is conducted separately for each model to ensure that there is no cross-model influence or learning.

Using these aligned scores, we fine-tune the baseline models through cRLHF, leveraging the Transformer Reinforcement Learning (TRL) framework \cite{vonwerra2022trl} to efficiently incorporate crowd-sourced feedback. The TRL framework allows us to fine-tune the models with LoRA by utilizing the language model's value head with PPO. Finally, we compare the performance of the fine-tuned models against their original versions, evaluating improvements across various tasks and datasets. 

\subsection{Human Feedback Collection}
To build the crowd-sourced human feedback necessary for our evaluation, we organized an annotation task structured as coursework assignments. A group of 10 selected students, serving as annotators, was recruited to evaluate 15 programming problem descriptions sourced from the LeetCode platform for each baseline model. To ensure there was no data overlap or leakage, we manually selected the problems from the dataset, ensuring that none of the selected tasks were included in the emulation benchmarks used for model evaluation. For each description, we generated 10 output samples from with three different levels: easy, medium and difficult, with the task of identifying and annotating errors in the generated code. Correct answers were predefined for these outputs to ensure accurate evaluation. The annotators represented varying levels of programming proficiency, ranging from beginners to experts, selected based on their previous coursework performance. 

During the experiment, the reliability of each annotator, denoted as $p_i$, was computed using the methodology described in Section \ref{sec:hacm}. The annotation tasks were conducted on an online platform designed for code evaluation, as depicted in Figure \ref{fig:ui}. Annotators were required to identify erroneous lines of code. As part of the setup, we followed the approach from \cite{ling2018human,rlhfbayesian0}, using the first 19 questions as ``honey pot'' questions to assess annotator reliability without revealing the correct answers. For each problem description, four distinct code samples were provided, and annotators were tasked with identifying any errors in the code.

\subsection{Dataset and Evaluation Metric}\label{sec:metric}
We evaluate the performance of pre-trained models specifically designed for text-to-code generation tasks. We utilize open-source pre-trained models that are trained on code as our baseline models, without any additional fine-tuning.  For benchmarks that include unit tests, such as HumanEval \cite{chen2021evaluating} and MBPP \cite{austin2021program}, we use the \texttt{Pass@k} metric to measure functional correctness. \footnote{\texttt{Pass@k} is an unbiased estimator for the probability that the correct answer is within $k$ samples, given a total of $n$ samples and $c$ unit tests. It is calculated as $\text{Pass@}k = \mathbb{E}\left[1 - \binom{n - c}{k} / \binom{n}{k}\right]$ \cite{chen2021evaluating}.} For each problem, we sample $n=100$ code solutions and evaluate $ k\in \{1, 10, 100\}$ of them with sampling temperatures $t = 0.8$. If any of these $k$ solutions pass all test cases, the problem is considered solved. To provide a comprehensive assessment of model performance, we report Pass@1, Pass@10, and Pass@100, which reflect the performance at different sample sizes.

\begin{table*}[h]
\centering
\caption{Evaluation results on the HumanEval and MBPP benchmarks. Each \texttt{Pass@k[\%]} (where $k \in {1, 10, 100}$) for each model is computed with the sampling temperature ($t = 0.8$). The table is divided into two parts for each benchmark: Baseline models (w/o) and models with cRLHF (w/). Each cell displays the pass rate percentage for a specific model, metric, and $k$ value, with the highest value in each metric highlighted.}
\setlength\extrarowheight{2.7pt}
\begin{tabular}{cc c ccc ccc}
\toprule
\multicolumn{1}{c}{\multirow{2}{*}{\textbf{Models}}} &\multicolumn{1}{c}{\multirow{2}{*}{\textbf{Parameters}}} &\multicolumn{1}{c}{\multirow{2}{*}{\textbf{cRLHF}}}& \multicolumn{3}{c}{\textbf{HumanEval}} & \multicolumn{3}{c}{\textbf{MBPP}} \\
\cmidrule(rl){4-6} \cmidrule(rl){7-9} 
\multirow{2}{*}{} & \multirow{2}{*}{} &\multirow{2}{*}{} &{Pass@1} & {Pass@10} & {Pass@100} &{Pass@1} & {Pass@10} & {Pass@100}\\
\midrule
\multirow{6}{*}{StarCoder2 \cite{lozhkov2024starcoder}} & \multirow{2}{*}{15B} &w/o& 25.6 & 43.3 & 50.6 & 35.6 & 69.5 & 84.2 \\ 
\multirow{6}{*}{}&&w/ & 25.6 & \textbf{44.1} & \textbf{53.7} & 35.7 & 69.6 & \textbf{84.6} \\
\addlinespace[0.1em]
\multirow{6}{*}{} & \multirow{2}{*}{7B} &w/o & 21.5 & 38.0 & 48.8 & 24.7 & 58.9 &76.8 \\
\multirow{6}{*}{}&&w/ & 21.5 & 38.4 & 48.8 & 24.2 & 58.8 &77.0 \\
\addlinespace[0.1em]
\multirow{6}{*}{} & \multirow{2}{*}{3B}&w/o & 20.2 & 35.3 & 43.3 & 21.9 & 55.0 &75.6 \\
\multirow{6}{*}{}&&w/ & 19.4 & 35.7 & 50.0 & 21.9 & 57.9 &76.0 \\
\addlinespace[0.5em] \hdashline[1pt/5pt]
\multirow{4}{*}{CodeLlama \cite{roziere2023code}} & \multirow{2}{*}{13B}&w/o & 20.1 & 35.1 & 45.1 & 22.7 & 58.4 &77.6 \\
\multirow{4}{*}{}&&w/ & 20.5 & 35.3 & 45.5 & 22.9 & 58.5 &77.8 \\
\addlinespace[0.1em]
\multirow{4}{*}{} & \multirow{2}{*}{7B}&w/o & 17.8 & 30.6& 39.6 & 25.0 & 56.4 &74.2 \\
\multirow{4}{*}{}&&w/ & 18.0 & 31.1 & 40.9 & 25.3 & 57.3 &74.4 \\
\addlinespace[0.5em] \hdashline[1pt/5pt]
\multirow{4}{*}{DeepSeek-Coder \cite{guo2024deepseek}} & \multirow{2}{*}{6.7B}&w/o & 22.8 & 35.6 & 42.7 & 37.6 & 70.9 &83.8 \\\addlinespace[0.1em]
\multirow{4}{*}{}&&w/ & 23.2 & 36.3 & 44.5 & 37.9 & \textbf{71.2} &84.0 \\
\addlinespace[0.1em]
\multirow{4}{*}{} & \multirow{2}{*}{1.3B}&w/o & 16.4 & 29.6 & 35.4 & 28.1 & 59.3 &77.4 \\\addlinespace[0.1em]
\multirow{4}{*}{}&&w/ & 16.8 & 30.1 & 39.0 & 28.6 & 59.8 &77.6 \\
\addlinespace[0.5em] \hdashline[1pt/5pt]
\multirow{4}{*}{InCoder \cite{fried2022incoder}} & \multirow{2}{*}{6B}&w/o & 7.0 & 20.7 & 36.6 & 7.0 &27.4 &52.6 \\
\multirow{4}{*}{}&&w/ & 8.6 & 20.9 & 36.6 & 8.7 & 32.8 &57.0 \\
\addlinespace[0.1em]
\multirow{4}{*}{} & \multirow{2}{*}{1B} &w/o & 5.8 & 15.0 & 24.4 & 3.5 & 18.5 &41.8 \\
\multirow{4}{*}{}&&w/ & 5.8 & 14.8 & 23.2 & 3.5& 18.7 &41.6 \\
\addlinespace[0.5em] \hdashline[1pt/5pt]

\multirow{4}{*}{CodeGemma \cite{team2024codegemma}} & \multirow{2}{*}{7B} &w/o & 7.2 & 26.2 & 38.4 & 20.0 & 55.2 &73.8 \\
\multirow{4}{*}{}&&w/ & 8.0 & 27.8 & 39.0 & 20.2 & 55.5 &73.8 \\
\addlinespace[0.1em]
\multirow{4}{*}{} & \multirow{2}{*}{2B} &w/o & 2.4 &8.8 & 17.1 & 0.9 & 5.9 &16.6 \\ 
\multirow{4}{*}{}&&w/ & 2.1 & 8.9 & 17.1 & 0.8 & 5.2 &15.0 \\

\addlinespace[0.5em] \hdashline[1pt/5pt]
\multirow{2}{*}{CodeQwen1.5 \cite{bai2023qwen}} & \multirow{2}{*}{7B} &w/o & 2.1 & 7.6 & 12.8 & 5.2 & 23.6 &47.2 \\
\multirow{2}{*}{}&&w/ & 2.1 & 7.1 & 14.6 & 5.0 & 22.7 & 47.0 \\
\addlinespace[0.5em] \hdashline[1pt/5pt]
\multirow{2}{*}{WizardCoder \cite{luo2023wizardcoder}} & \multirow{2}{*}{15B} &w/o & 30.1 & 41.5 & 49.4 & \textbf{40.8} & 62.7 & 74.4 \\
\multirow{2}{*}{}&&w/ & \textbf{30.3} & 42.7 & 49.4 & \textbf{40.8 }& 62.8 &74.6 \\
\addlinespace[0.5em] \hdashline[1pt/5pt]
\multirow{2}{*}{CodeGen2.5 \cite{nijkamp2023codegen2}} & \multirow{2}{*}{7B} &w/o &21.1 & 36.2 & 46.3 & 23.1 & 55.4 &70.4 \\
\multirow{2}{*}{}&&w/ & 21.2 & 35.8 & 45.7 & 23.7 & 57.0 &75.6 \\
\addlinespace[0.5em]\hdashline[1pt/5pt]
\multirow{6}{*}{PolyCoder \cite{xu2022systematic}} & \multirow{2}{*}{2.7B}&w/o & 3.0& 7.8 & 13.4 & 0.5 & 3.8 &18.2 \\
\multirow{6}{*}{}&&w/ & 2.9 & 7.5 & 14.0 & 0.5 & 3.8 &15.8 \\
\addlinespace[0.1em]
\multirow{6}{*}{} & \multirow{2}{*}{0.4B}&w/o & 1.3 & 3.9 & 8.5 & 0.06 & 0.6 &4.0 \\
\multirow{6}{*}{}&&w/ & 1.3 & 4.2 & 9.1 & 0.07 & 0.6 &4.2 \\
\multirow{6}{*}{} & \multirow{2}{*}{160M} &w/o & 1.0 & 3.0 & 4.9 & 0.04 & 0.3 &2.4 \\
\multirow{6}{*}{}&&w/ &1.0 & 2.8 & 4.9 & 0.04 & 0.3 &2.8 \\
\bottomrule
\end{tabular}

\label{table:results}
\end{table*}

\subsection{Baseline Models}
We assess code generation performance using a diverse set of public pre-trained language models, ranging up to 15 billion parameters to ensure comprehensive analysis. Only models with a value head in their architecture, essential for our framework, are included as baselines.

Our evaluation set includes models such as WizardCoder-15B \cite{luo2023wizardcoder}, StarCoder2-15B/7B \cite{lozhkov2024starcoder}, and CodeLlama-13B/7B \cite{roziere2023code}. Additionally, we assess CodeGen2.5-7B \cite{nijkamp2023codegen2}, CodeGemma-7B/2B \cite{team2024codegemma}, and CodeQwen1.5-7B \cite{bai2023qwen}. Our set also features DeepSeek-Coder-6.7B/ 1.3B \cite{guo2024deepseek} and InCoder-6B/1B \cite{fried2022incoder}. For smaller-scale models, we evaluate PolyCoder-2.7B/0.4B/100M \cite{xu2022systematic}. This diverse selection allows us to assess our framework's performance across a broad spectrum of model sizes and architectures. Although each model has reported its performance on existing benchmarks, we reproduce all experiments under our specified settings, as outlined in Section \ref{sec:setup}, to ensure a fair comparison. For example, some models, such as WizardCoder, may use greedy decoding for optimal performance in their original setup, or models like InCoder may require a language-specific prefix before the prompt to provide contextual hints. In our baseline setup, however, we apply a standardized configuration, which may result in lower performance, but ensures consistency across evaluations.

\begin{figure*}
	\centering
         \begin{subfigure}{0.49\textwidth}
		\includegraphics[width=\linewidth]{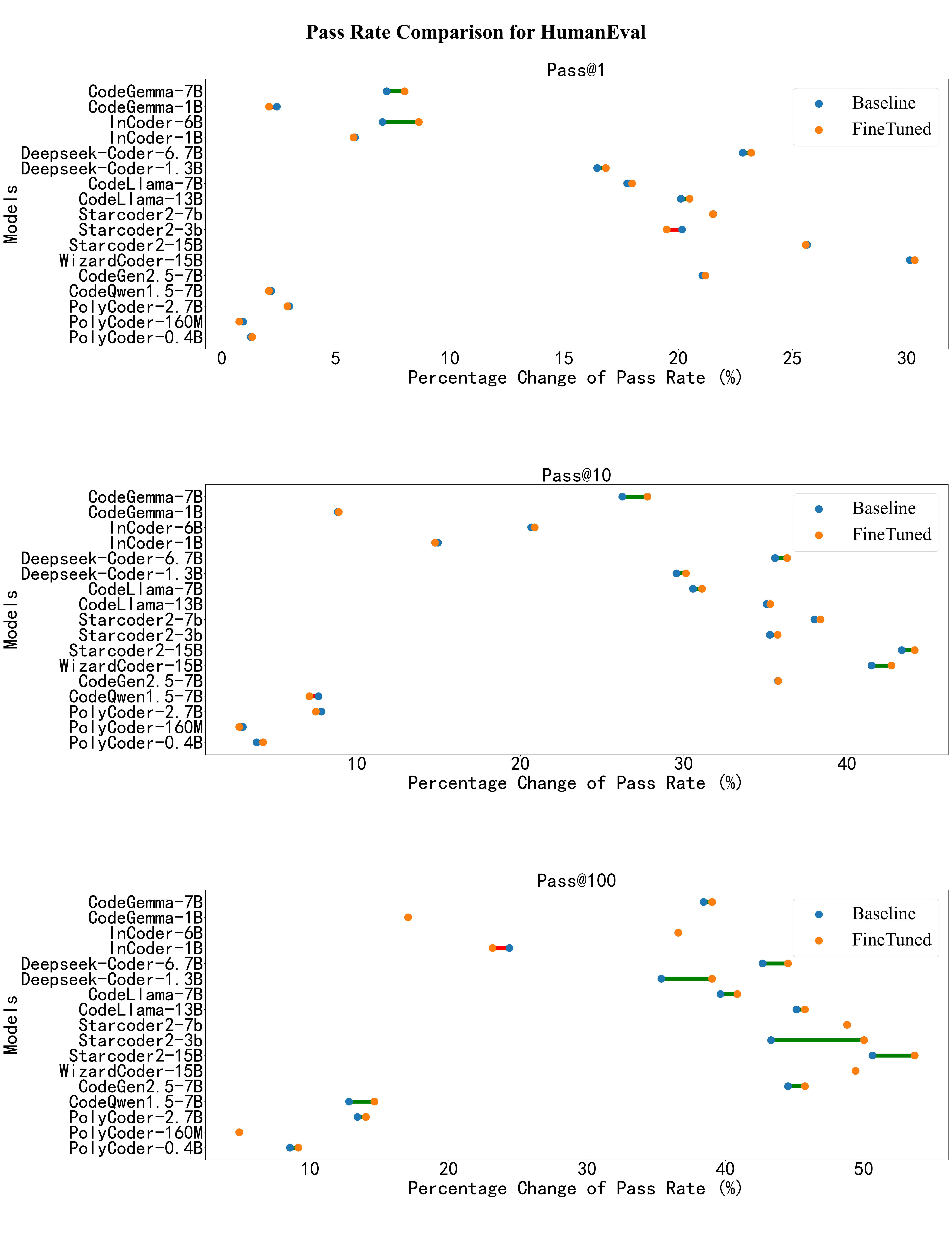}
		\captionsetup[subfigure]{justification=centering}
		\label{fig:pass_1}
	\end{subfigure}
        \begin{subfigure}{0.49\textwidth}
		\includegraphics[width=\linewidth]{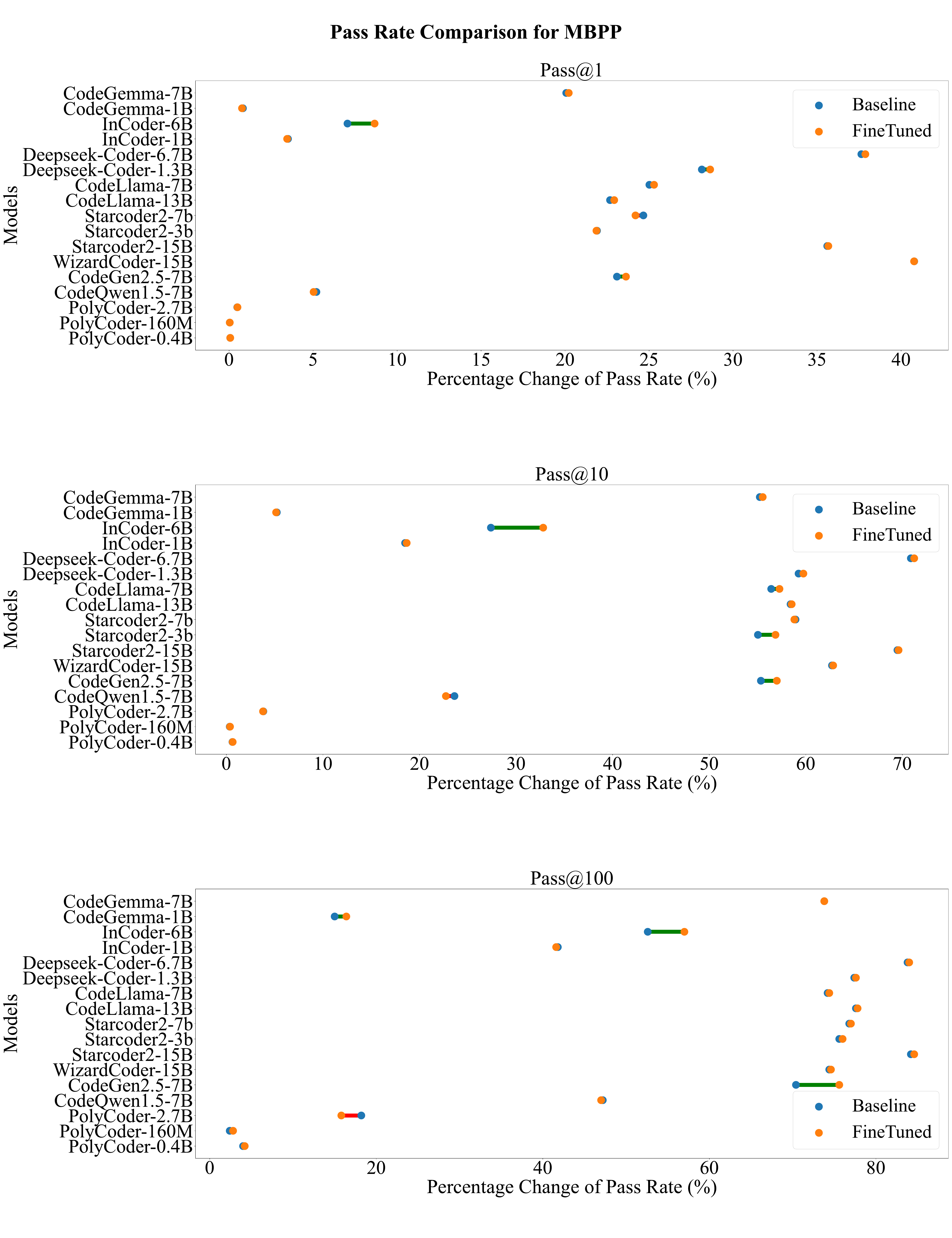}
		\captionsetup[subfigure]{justification=centering}
		\label{fig:pass_2}
	\end{subfigure}
	\caption{Comparative performance analysis of the baseline and fine-tuned models on the HumaEval task (left) and the MBPP task (right). Each point represents the performance at a specific model, with lines connecting the baseline and fine-tuned results. Green lines indicate an improvement in performance after fine-tuning, while red lines indicate a decrease}
	\label{fig:passrate_results}
\end{figure*}

\section{Results and Discussion} \label{sec:results}
\subsection{Numerical Results}
The main numerical results from our experiments with different models are showcased in Table \ref{table:results}. This table provides a summary of the outcomes obtained from the HumanEval and MBPP benchmarks for both the original LLM and the model incorporating the proposed cRLHF.  In the baseline results for HumanEval, we see that larger models generally perform better across all three metrics. For example, the largest model, WizardCoder-15B, achieves the highest Pass@1 score of 30.1\%, followed by StarCoder2-15B at 25.6\%. As expected, smaller models such as PolyCoder-160M and PolyCoder-0.4B demonstrate significantly lower performance. There are noticeable performance differences, highlighting the varying effectiveness of different architectures and training data even though the number of parameters is similar. 

After fine-tuning the baseline models with cRLHF, a modest improvement in several cases can be observed. For instance, StarCoder2-15B has the best performance across all the models in Pass@10 and Pass@100 and sees a slight increase in its Pass@10 score from 43.3\% to 44.1\%, while its Pass@100 improves from 50.6\% to 53.7\%. Similarly, the performance of WizardCoder-15B remains relatively stable with little fluctuations, but performs the best on the Pass@1, indicating that fine-tuning can enhance its performance.

For the MBPP task, models without cRLHF, such as WizardCoder-15B and DeepSeek-Coder-6.7B, show superior performance, with Pass@1 scores of 40.78\% and 37.6\%, respectively. DeepSeek-Coder-6.7B also leads in Pass@10 metrics, highlighting their robustness in code generation across multiple output ranks. StarCoder2-15B achieves the highest performance in the Pass@100 metrics. In contrast, smaller models like PolyCoder-0.4B and PolyCoder-160M show lower performance and minimal improvements in Pass@10 and Pass@100. The fine-tuning process using cRLHF has led to observable improvements in model performance, though the extent of these improvements varies across different models. Notably, the StarCoder2-15B model demonstrates a modest increase in Pass@100, rising from 84.2\% to 84.6\%. This suggests that cRLHF can enhance performance, especially in larger models. Remarkably, the DeepSeek-Coder-6.7B, despite having relatively fewer parameters, still achieves over 84.0\% in Pass@100. Smaller models also benefit from cRLHF, though less dramatically. For example, CodeLlama-7B shows an improvement in Pass@1 from 25.0\% to 25.3\% and Pass@100 from 74.2\% to 74.4\%. However, some models, such as PolyCoder-160M, show little changes in performance after fine-tuning, reflecting the limited impact of cRLHF on models with fewer parameters.

Overall, fine-tuning with cRLHF generally leads to performance improvements, particularly for larger models. However, the extent of these improvements varies across different models and metrics. The changes in pass rates for different models are illustrated in Figure \ref{fig:passrate_results}. In the HumanEval task, out of the 17 models evaluated, 8 models showed improvement in the Pass@1 score, 12 models in the Pass@10 score, and 10 models in the Pass@100 score after fine-tuning. In the MBPP task, 9 models showed improvement in Pass@1, while 10 models showed improvement in both Pass@10 and Pass@100. These results suggest that fine-tuning generally had a positive impact on model performance in both tasks, with more substantial gains observed in Pass@10 and Pass@100 scores. This also indicates that fine-tuning had a positive impact on the performance of the models in both tasks, with a more pronounced effect on the Pass@10 and Pass@100 scores.

To ensure that improvements in Pass@k metrics were not due to random variations, we have grouped the models by parameter size and analyzed the results for each group. Overall, for the HumanEval benchmark, the improvements in Pass@1, Pass@10, and Pass@100 metrics were 0.2\%, 0.3\%, and 1.2\%, respectively. In contrast, for the MBPP benchmark, the improvements were 0.2\%, 0.6\%, and 0.6\%. This analysis revealed notable changes in performance metrics after applying our cRLHF fine-tuning approach.

For smaller models with fewer than 6 billion parameters, we have observed a 0.1 improvement in Pass@1, 0.12 in Pass@10, and 1.3 in Pass@100 on HumanEval, with slightly higher gains for MBPP, including a 0.25 increase in Pass@1 and nearly 1 in Pass@10. In the 6-10 billion parameter range, models showed more balanced improvements, with HumanEval seeing a 0.23 gain in Pass@1 and 0.44 in Pass@10, while MBPP showed a 0.1 rise in Pass@1 and a 1 jump in Pass@100. Larger models, exceeding 10 billion parameters, displayed steady improvements, with HumanEval seeing a 0.74 gain in Pass@10 and MBPP showing a more moderate 0.27 increase in Pass@100. By analyzing grouped results across different model sizes, we can better attribute improvements to model enhancements rather than randomness. Despite its small scale, our experiment exhibited significant performance improvements across all model sizes in code generation, highlighting the effectiveness of cRLHF even with limited resources.

\subsection{Discussion}
Our findings demonstrate the significant impact of cRLHF on improving code generation quality. While the improvements may be small, they are particularly significant in models with larger parameter sizes, suggesting that cRLHF has the potential to enhance code generation, especially when applied to more complex models. The consistent improvements observed across various benchmarks show that embedding cRLHF into existing models can effectively enhance both the relevance and quality of generated code. These advancements are particularly promising for speeding up software development and AI-assisted programming tasks. Exploring alternatives, such as different baselines, fine-tuning datasets, and learning rate algorithms, further broadens the potential applications and highlights how our framework can adapt to various use cases within the domain.

In more general cases with multiple annotators and varying noise levels, the regularization term can be adjusted to account for factors like annotator confidence or entropy-based regularization to balance uncertainty in $p_i$. This enhances model robustness when annotator quality varies. Proximal algorithms systematically ensure stability while refining predictions and can be extended to handle complex scenarios, including different noise models or hierarchical annotator structures.

However, the different setups of LLMs, such as variations in architecture, parameter size, and the datasets used for pre-training or fine-tuning, can lead to varying results that might not apply to all situations. Additionally, the scale of the human feedback data in our experiments is relatively small, which could limit the observable improvements. Still, our findings offer valuable insights into connecting human-in-the-loop systems with modern LLMs. This paper also highlights how using crowdsourced human feedback can make the computational process clearer and improve the interaction between human expertise and advanced language models.

\subsection{Limitations}
Although we have seen improvements, the codes generated by LLMs and annotated by humans may still not be fully accurate or optimal. Also, human feedback used to guide the RL process can sometimes introduce errors or noise. A significant limitation of this approach is the small scale of the human feedback, which can limit the extent of observable improvements. Enlarging the human feedback dataset is crucial, but it poses challenges, particularly in ensuring the diversity and quality of feedback. It can also be difficult to find sufficient numbers of skilled annotators for large-scale human feedback collection. As a possible solution for scaling up feedback collection, future work could explore the LLM-as-a-Judge for AI feedback \cite{bai2022constitutional,tan2023large} to assess and validate generated code, enabling the use of larger, more diverse datasets while maintaining consistent feedback quality. Also, scalability problems in crowd-sourced frameworks often become a significant concern, as it can be challenging to find suitable datasets and workers for the sampling process. 

Another limitation of our approach is its reliance on causal language models with a value head, which necessitates applying TRL. In addition, this setup can be computationally intensive, as fully fine-tuning the model requires substantial GPU resources. Generating a large number of samples for training also involves significant computational demands. These factors can limit the scalability of our method and may require considerable investment in hardware and recruiting annotators to achieve optimal results.

\subsection{Extensions to Domain-specific Language}

By leveraging the software naturalness hypothesis, we can extend our cRLHF framework to encompass other domain-specific languages. One notable example is specialized convex optimization modeling languages, such as CVX \cite{cvx,cvxpy} that allow users to model and formulate optimization problems rapidly. Its code generation counterpart (e.g., CVXGEN in \cite{mattingley2012cvxgen}, CVYPyGen in \cite{schaller2022}) then allow users to derive practical programming language from these high-level domain-specific language. For example, CVXGen and CVXPyGEN converts, respectively, Python and Matlab code into optimized C code with embedded custom solvers optimized for specific convex optimization problem structures, resulting in lightweight and memory-efficient codebase that is well-suited for resource-constrained environments. The use of large language models thus allow the problem complexity to be decomposed into different stages of optimization in code generation, making them effective for large-scale and high-dimensional problems.

\section{Conclusion}\label{sec:conclusion}
In summary, our paper introduces cRLHF, an innovative framework that combines RL with human-assisted computation to enhance code generation in LLMs. By integrating RLHF with Bayesian inference, our approach aligns rankings from multiple annotators and computes reward scores without additional reward modeling. This novel method not only advances AI-assisted programming but also provides developers with a powerful tool for creating complex conversational agents and improving software development processes. Our evaluation of the cRLHF framework on existing benchmarks, with real-world annotators assessing generated outputs, demonstrated that it improves performance compared to recent baselines in code generation tasks.

The principles behind the cRLHF framework have broader applications beyond code generation. By integrating human feedback with reinforcement learning, this approach can extend to fields like mathematical reasoning and logical deduction, enhancing model interpretability and performance through alignment with human expertise. This opens new opportunities for transforming how AI systems interact with and learn from human input across various domains. Future work include integrating the Bayesian optimization framework with domain-specific languages to improve text-to-code generation and enhance the quality of crowd-sourced human feedback through automated tools. Addressing program modification during execution in line with Dijkstra's principles \cite{dijkstra}, expanding RLHF applications to debugging and code review, and exploring multi-modal feedback sources will enrich the training process. Additionally, conducting longitudinal studies on the impacts of AI-assisted programming and creating collaborative coding environments will foster innovation and improve developer experience while ensuring fairness and inclusivity in AI tools.
\newpage
\bibliographystyle{IEEEtran}
\bibliography{reference}
\end{document}